\documentclass[conference]{IEEEtran}
\IEEEoverridecommandlockouts
\usepackage{cite}
\usepackage{amsmath,amssymb,amsfonts}
\usepackage{graphicx}
\usepackage{textcomp}
\usepackage{xcolor}
\def\BibTeX{{\rm B\kern-.05em{\sc i\kern-.025em b}\kern-.08em
    T\kern-.1667em\lower.7ex\hbox{E}\kern-.125emX}}

\usepackage{amsthm}

\newtheorem{definition}{Definition}
\usepackage{float}
\usepackage{booktabs}
\usepackage{algorithm}
\usepackage{algpseudocode}

\usepackage{xurl}

\DeclareMathOperator*{\argmin}{arg\,min}

\begin{document}

\title{Provably Safe Model Updates}

\makeatletter
\newcommand{\linebreakand}{%
  \end{@IEEEauthorhalign}
  \hfill\mbox{}\par
  \mbox{}\hfill\begin{@IEEEauthorhalign}
}
\makeatother

\author{

\IEEEauthorblockN{Leo Elmecker-Plakolm$^\star$}
\IEEEauthorblockA{\textit{Department of Computing}\\
\textit{Imperial College London}\\
London, United Kingdom\\ \texttt{le24@ic.ac.uk}}
\and
\IEEEauthorblockN{Pierre Fasterling$^\star$}
\IEEEauthorblockA{\textit{School of Computer and Comm. Sciences (IC)}\\
\textit{École Polytechnique Fédérale de Lausanne}\\
Lausanne, Switzerland\\ \texttt{pierre.fasterling@epfl.ch}}
\and
\IEEEauthorblockN{Philip Sosnin$^\star$}
\IEEEauthorblockA{\textit{Department of Computing}\\
\textit{Imperial College London}\\
London, United Kingdom\\ \texttt{p.sosnin23@imperial.ac.uk}}
\linebreakand 
\IEEEauthorblockN{Calvin Tsay}
\IEEEauthorblockA{\textit{Department of Computing}\\
\textit{Imperial College London}\\
London, United Kingdom\\ \texttt{c.tsay@imperial.ac.uk}}
\and
\IEEEauthorblockN{Matthew Wicker}
\IEEEauthorblockA{\textit{Department of Computing}\\
\textit{Imperial College London}\\
London, United Kingdom\\ \texttt{m.wicker@imperial.ac.uk}}
}

\maketitle

\begin{abstract}

Safety-critical environments are inherently dynamic. Distribution shifts, emerging vulnerabilities, and evolving requirements demand continuous updates to machine learning models. Yet even benign parameter updates can have unintended consequences, such as catastrophic forgetting in classical models or alignment drift in foundation models. Existing heuristic approaches (e.g., regularization, parameter isolation) can mitigate these effects but cannot certify that updated models continue to satisfy required performance specifications. We address this problem by introducing a framework for provably safe model updates. Our approach first formalizes the problem as computing the largest locally invariant domain (LID): a connected region in parameter space where all points are certified to satisfy a given specification. While exact maximal LID computation is intractable, we show that relaxing the problem to parameterized abstract domains (orthotopes, zonotopes) yields a tractable primal-dual formulation. This enables efficient certification of updates—independent of the data or algorithm used—by projecting them onto the safe domain. Our formulation further allows computation of multiple approximately optimal LIDs, incorporation of regularization-inspired biases, and use of look-ahead data buffers. Across continual learning and foundation model fine-tuning benchmarks, our method matches or exceeds heuristic baselines for avoiding forgetting while providing formal safety \looseness=-1 guarantees.


\end{abstract}

\section{Introduction}

\begin{figure}
    \centering
    \includegraphics[width=\linewidth]{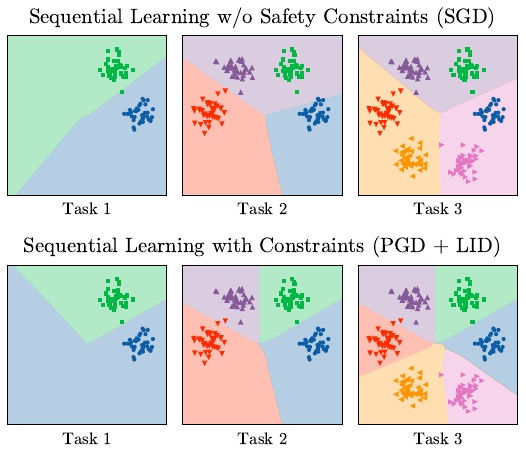}
    \caption{Illustration of our method on a simple `blobs' dataset, where the model sequentially learns pairs of classes. The true class label for each data point corresponds to the color of the point and the background region to the output of the model. Top: Standard training forgets previously learned classes. Bottom: Training within our locally invariant domain successfully preserves performance on earlier tasks.}
    \label{fig:blobs_example_figure}
\end{figure}

Machine learning capabilities are evolving at unprecedented speed, driving deployment in high-stakes domains where safety and security are critical requirements \cite{bommasani2021opportunities, szpruch2025insuring}. Models used in such critical domains (e.g., autonomous driving, medical applications, and financial services) must continually contend with input distribution shifts, newly discovered vulnerabilities, and evolving regulatory requirements. As a result, regular updates are a prerequisite to safe deployment.
However, even benign updates can trigger severe unintended effects on model behavior—most notably catastrophic forgetting and loss of alignment—which are extensively documented in both foundation models \cite{qi2023fine, yang2023shadow, zhan2023removing, lermen2023lora} and in continual learning \cite{kirkpatrick2017overcoming, wang2024comprehensive}. In absence of \textit{a priori} guarantees, model updates cannot be deployed into safety critical contexts without complete re-evaluation to satisfy regulators \cite{us2024predetermined} or internal auditors \cite{mokander2024auditing, wicker2025move}. Yet comprehensive re-validation is often impractical due to prohibitive computational cost, latency constraints, or inaccessible training data. This tension between the need for adaptation and the demand for assurance is a central obstacle to deploying learning systems safely in the real world.

Prior works usually seek to combat catastrophic forgetting or preserve model alignment during updates through approaches such as regularization \cite{li2017learning, shen2020towards}, replay buffers \cite{rebuffi2017icarl, huang2024lisa}, or parameter isolation \cite{rusu2016progressive, kang2022forget}. While these methods display promising results, they do not come with formal guarantees of avoiding forgetting and therefore cannot avoid the need for revalidation. To address this, we investigate the problem of providing an update framework that has \textit{a priori} guarantees that the updated model satisfies a given performance specification e.g., prior task performance or predictive stability.  Beyond eliminating the revalidation bottleneck, we discuss how such a framework may offer certifiable safety evidence for regulators \cite{us2024predetermined}, reliable model version control, enforceable terms-of-use for downstream users \cite{anderljung2023frontier, qi2024evaluating}, and safe mechanisms for adapting to emerging vulnerabilities.

To provide safe update guarantees, we introduce the problem of computing a local invariant domain (LID): a connected region of parameter space within which all parameters are provably certified to satisfy a specified performance requirement measured on the prior task. Once a LID has been computed, the safety of an arbitrary update can be efficiently enforced by projecting any proposed parameter change onto the certified LID, and where multiple tasks are observed in sequence one can project onto the intersection of each task's LID to certify safety (discussed in detail in Section~\ref{sec:methods}).

The primary technical contribution of this paper is the development of effective and practical LID computation techniques as not all LIDs are practically useful. For example, the LID containing only the initial model parameter vector is trivially valid, but effectively prohibits model updates. Thus, we formulate the general LID problem as finding the \textit{maximal} such domain (by volume) over all connected subsets of the parameter space. Because optimizing over arbitrary connected domains is intractable, we propose a tractable yet conservative relaxation: an optimization problem over parameterized abstract domains, which allows us to leverage advances in bounding neural networks \cite{gowal2018effectiveness, tsay2021partition} and abstract interpretation \cite{gehr2018ai2, sosnin2024certified}. Of course, the optimization problem over parameterized abstract domains must be constrained to only domains for which we can certify that every parameter respects the safety property. Thus, we instantiate a primal-dual formulation of the constrained maximal LID problem enabling the computation of approximately maximal LIDs via alternating gradient-descent-ascent. 

Practical implementation of this framework requires careful solutions to several important problems to avoid unsound or trivial LIDs. Firstly, where safety constraints are formulated as properties of an input distribution, one must use a finite-sample approximation of the constraint thus introducing a probability of failure. To address this, we provide concentration bounds that allow us to give worst-case probabilistic guarantees of constraint satisfaction. Secondly, our primal-dual formulation only finds locally optimal solutions to the LID problem. To address this, we propose to use prior work from continual learning to bias our optimization towards better solutions and propose to compute multiple LIDs (LID multiplicity) to avoid problematic local optima. Finally, many model updating scenarios either maintain access to previous task data or have prior information about what future updates might be relevant, e.g., planned support for additional disease distributions in medical applications or planned multi-lingual support for language models. Taking inspiration from the continual learning community, we support such scenarios by introducing a finite-buffer extension to the LID problem.

In Section~\ref{sec:experimental}, we experimentally demonstrate the effectiveness of our approach across standard continual learning and foundation model fine-tuning tasks. We benchmark the performance of our algorithm against established regularization-based approaches and benchmark the guarantees of our algorithm against the continual learning approach \textit{InterContiNet} (ICN) \cite{wolczyk2022continual}. An embedded step of the ICN algorithm can be viewed as the computation of a LID, though the authors do not pose the problem as such. Thus, ICN does not consider finite sample error, use of replay buffers, LID multiplicity, or principled constraint encoding which leads to potentially trivial or unsound bounds and numerical instability. Across a wide variety of synthetic and real-world benchmarks including MNIST \cite{lecun1998gradient}, CIFAR-10 \cite{CIFAR}, hate speech classification, and multi-lingual adaptation, we find that our approach is able to match or best benchmark performance while providing state-of-the-art guarantees. In summary, this paper makes the following contributions: 

\begin{itemize}
    \item We introduce the \emph{maximum local invariant domain (LID)} problem, and develop a general framework for computing provable guarantees that model updates preserve behavioral specifications such as alignment or task \looseness=-1 performance.
    \item We demonstrate how LIDs can be used to certify consecutive tasks and can leverage replay buffers thus leading to a more general approach for certified continual learning.
    \item We propose a relaxation of the maximum LID via abstract interpretation techniques, enabling an efficient first-order primal-dual algorithm to approximate maximal LIDs.
    \item We extend our LID computation algorithm using biasing methods inspired by continual learning literature to identifying empirically favorable LID solutions.
    \item We evaluate our approach across synthetic and real-world benchmarks, including MNIST, CIFAR, and content moderation, demonstrating its ability to provide non-trivial and practically useful safety guarantees for model updates.
\end{itemize}

\section{Related Works}

\paragraph{Model Updates Eroding Alignment\protect\footnote{We provide an extended discussion of related works in Appendix~\ref{app:related-works}.}} Ensuring alignment of foundation models is a growing area of research \cite{huang2024harmful, zhaosurvey, huang2024survey}. Despite extensive training-time safeguards, it has been shown that fine-tuning rapidly erodes many safeguards \cite{yang2023shadow}, even when users are not malicious \cite{qi2023fine}. Some heuristics have been developed to defend against this erosion \cite{mukhoti2023fine, wei2024assessing, rosati2024representation}, but have been found unsuccessful in the face of more sophisticated attackers \cite{qi2024evaluating, rando2025adversarial}. These studies indicate the start of an arms race between attackers and defenders often observed in adversarial settings, and they point to the importance of formal verification in safety-critical contexts.

\paragraph{Catastrophic Forgetting in Continual Learning} Ensuring that training updates do not degrade prior model capabilities (as seen in Figure~\ref{fig:blobs_example_figure}), an issue termed catastrophic forgetting \cite{kirkpatrick2017overcoming, li2017learning, van2022three}, has been widely studied within the continual learning community \cite{wang2024comprehensive}. Avoiding forgetting is often studied under the assumption that data from prior tasks are unavailable \cite{van2022three}. In this case, heuristic forgetting defenses can largely be split into approaches that use parameter isolation (i.e., only updating a subset of parameters \cite{rusu2016progressive, yoon2017lifelong, kang2022forget}) or regularization \cite{li2017learning, kirkpatrick2017overcoming, shen2020towards} (i.e., soft constraints limiting changes to important parameters). With access to prior task data for revalidation, replay methods have been developed \cite{rebuffi2017icarl, huang2024lisa}. Each of these methods has demonstrated some considerable success in reducing forgetting; however, they do not come with concrete, formal guarantees.

\looseness=-1
\paragraph{Formal Methods in Machine Learning}
The field of formal methods for machine learning algorithms has produced proofs of a wide range of notions including adversarial robustness \cite{gowal2018effectiveness, sosnin2024certified}, explainability \cite{wicker2022robust}, privacy \cite{wicker2024certification}, and uncertainty \cite{bitterwolf2020certifiably}. Most related to this paper is the growing literature on formal security guarantees for machine learning models \cite{wang2022improved, sosnin2024certified} which, however, focus on specific dataset threat models.
The most closely related work is \textit{InterContiNet} (ICN) \cite{wolczyk2022continual}, which introduces a continual learning algorithm that uses heuristic optimization to identify safe parameter intervals.
The algorithm begins with a large, potentially unsafe parameter interval and iteratively contracts it via SGD to meet an accuracy threshold.
In contrast to our approach, ICN seeks only to find \textit{some} valid LID rather than a \textit{maximal} LID, reducing the model’s flexibility to learn new tasks while preserving safety.
Moreover, its shrinkage-based optimization procedure is susceptible to interval collapse and numerical instability, which hinders scalability to larger models.
Finally, the method does not incorporate multiplicity, biasing, replay buffers, or finite-sample safety guarantees---key contributions of this work.
\section{Certification for Provably Safe Updates}\label{sec:methods}

Throughout the paper we will consider tasks to be supervised learning problem with a fixed sample size $N$, i.e., $\mathcal{D} := \{ (x^{(i)}, y^{(i)}) \}_{i=1}^{N}$, where we describe $x \in \mathbb{R}^{n}$ as our feature vectors and $y \in \mathcal{Y}$ to be our labels; we will assume labels inhabit a $c$-dimensional space (i.e., $c$ classes or $c$-dimensional regression). We denote machine learning models as parameterized functions $f^{\theta} : \mathbb{R}^n \to \mathcal{Y}$ with parameters $\theta \in \mathbb{R}^p$. A proposed update to the parameters comprises a vector $u \in \mathbb{R}^{p}$, with the result of applying the update being: $\theta' = \theta + u$. We denote a function that proposes model updates (SGD, SFT, DPO, etc.) as an update mechanism $\mathcal{M}$. Finally, we model behavioral properties of interest such as alignment or performance criteria using a function $\phi : \mathbb{R}^{p \times n \times c} \to \mathbb{R}$. Without loss of generality, we use the convention that lower scores of $\phi$ are better and indicate better performance/alignment, and in our experiments we take $\phi$ to be one minus the accuracy on a particular task, but this is left purposefully generic with further discussion of concrete choices in Appendix~\ref{app:heuristics}, alongside a collection of notation used throughout this section in Appendix~\ref{subsec:method_notation}.
We now provide the general problem statement for a provably safe update mechanism: 
\begin{definition}\label{def:provable-update} ($\delta$-Safe Update Mechanism) Given a machine learning model $f^{\theta}$ trained on a task with data distribution $P(X,Y)$ and a property $\phi$ we say that a mechanism $\mathcal{M}$ taking in an arbitrary dataset $\mathcal{D}'$ and parameter $\theta$ is a provably $\delta$-safe update mechanism if 
\begin{equation}
    \forall \mathcal{D}', \quad \mathbb{E}_{P(X,Y)}\Big[ \phi\big(\theta + \mathcal{M}(\theta, \mathcal{D}'), x, y\big)\Big] \leq \delta.
\end{equation}
\end{definition}
Before proceeding to discuss how one might formulate a non-trivial mechanism satisfying this definition, we highlight several key implications of the definition. First, the definition is independent of the updated model's performance on the new task (presumably represented by $\mathcal{D}'$), as this would be unrealistic given that $\mathcal{D}'$ is unspecified and thus may be adversarially chosen. Second, the fact that it is possible to choose $\mathcal{D}' = \emptyset$ implies that the existence of any $\delta$-safe update mechanism depends upon the initial parameter being satisfactory, i.e.,  $\mathbb{E}_{P(X,Y)}\big[ \phi\big(\theta, x, y\big)\big] \leq \delta$. Finally, given that this inequality holds, we highlight that the trivially safe update mechanism is the do-nothing mechanism, i.e., $\forall \mathcal{D}', \mathcal{M}(\theta, \mathcal{D}') = \mathbf{0}$ which satisfies the definition.

\subsection{Local Invariant Domains (LIDs)}

As a foundation for constructing provably sound update mechanisms, we now introduce the notion of Local Invariant Domains (LIDs).
\begin{definition}\label{def:LID} (Locally Invariant Domain) Given a machine learning model $f^{\theta}$ trained on a task with data distribution $P(X,Y)$, and a property $\phi$, we say that a connected domain $T \subseteq \mathbb{R}^{p}$ containing $\theta$ is a provably $\delta$-invariant domain if
$$\forall \theta' \in T, \, \mathbb{E}_{P(X,Y)}\Big[ \phi\big(\theta', x, y\big)\Big] \leq \delta.$$
\end{definition}
Any update mechanism restricted to outputs in a domain satisfying Definition~\ref{def:LID} is one that satisfies Definition~\ref{def:provable-update}; however, it may be that $T = \{\theta\}$ which yields the trivial mechanism. To learn non-trivial safe update mechanisms, where possible, we propose a stronger, modified definition: 
\begin{definition}\label{def:MLID} (Maximal Locally Invariant Domain) Given a machine learning model $f^{\theta}$ trained on a task with data distribution $P(X,Y)$ and a property $\phi$, we say that a connected domain $T \subseteq \mathbb{R}^{p}$ containing $\theta$ is a \textit{maximally} provably $\delta$-invariant domain with respect to a specified size metric $|\cdot|_S$ (e.g., Hausdorff measure), if it is a solution to the following optimization problem:  
\begin{equation}
\max_{T} |T|_{S} \, \, \mathrm{s.t.} \, \, \forall \theta' \in T, \, \mathbb{E}_{P(X,Y)}\Big[ \phi\big(\theta', x, y\big)\Big] \leq \delta.\label{eq:mlid}
\end{equation}
\end{definition}
\newpage
\subsection{Provably Safe Updates via LIDs}
In this section, we connect the notions of safe update mechanisms and LIDs, highlighting specific cases of how LIDs can be used to augment popular model updating mechanisms.

\subsubsection{One-Step Updates (Fine-Tuning)} Given access to a known LID, one can make any arbitrary update mechanism $\mathcal{M}$ a provably safe mechanism by transforming it into $M_{T} := \Pi_{T}(M(\theta, \mathcal{D}'))$ where $\Pi$ is the standard projection operator $\Pi_T(\theta) := \argmin_{\theta^\star \in \bar{T}} ||\theta - \theta^\star||$ and $\bar{T}$ is the closure of $T$ to ensure this is well-defined. We emphasize that while the projection itself can be a nontrivial operation, any approximate projector that results in a parameter that belongs to $T$ is satisfactory.
While projection satisfies the definition, we can go further by carefully choosing the best $T$. To guide selection, we observe that for any loss function $\mathcal{L}$ that one wishes to minimize for the new task $\mathcal{D}'$, observe that $\min_{\theta \in T^{\star} }\mathcal{L}(\theta, \mathcal{D}') \leq \min_{\theta \in T }\mathcal{L}(\theta, \mathcal{D}')$ if $T \subset T^\star$. 
Thus, intuitively, computing the largest domain $T$ yields the best performance on subsequent downstream tasks. We highlight that purely finding the \textit{largest} domain may not be effective on its own, as the set may only be large in directions that are not helpful for improving downstream performance. Research in sparse updates in parameter isolation \cite{rusu2016progressive} or efficient fine-tuning \cite{wei2024assessing} may represent powerful heuristics for identifying important directions.  Additionally with access to some small portion of data from the desired downstream task, one can estimate the important directions. We discuss both of these approaches for improving  LID computation in Section~\ref{para:biasingLIDs}.

\subsubsection{Multi-Step Updates (Continual Learning)} In continual learning, the model is updated multiple times~\cite{van2022three}, and each update must ensure that all previous tasks respect a given constraint. Definition~\ref{def:provable-update} can be generalized to $k$-many, ordered updates by defining the $k^{th}$ update safety as: $\forall \mathcal{D}'_k, \forall i \in [k-1],$ 
\begin{align}
&\mathbb{E}_{P(X_{i},Y_{i})}\Big[ \phi\big(\theta + \sum_{j=1}^{k-1}\mathcal{M}(\theta^{(j-1)}, \mathcal{D}'_j), x, y\big)\Big] \leq \delta,
\end{align}
where $\theta^{(j)}$ is the result of the first $j$ updates and the initial $\theta' = \theta^{(0)}$. Though the definition of safety has changed, the fundamental solution with LIDs needs only a minor modification to generalize to this case. For the $i^{th}$ (previous) task, we compute a LID $T_i$, and we can then compute a new LID that holds for all tasks as $T^\star = \cap_{j=1}^{i} T_j$, the intersection of all independently computed LIDs. We highlight that this intersection does not need to be explicitly/analytically computed as we only require projection onto the intersection which can be done approximately.

\subsubsection{Use of Replay Buffer}
An important assumption we make in the formulation of our optimization problem (Definition~\ref{def:MLID}) is that the LID is optimized over only one joint distribution $P(X,Y)$. This encodes the assumption that after the model is trained on a task, the corresponding test data can no longer be accessed. Many popular approaches to combating forgetting relax this assumption by allowing a small amount of replay data. Allowing access to data from prior tasks, we can jointly optimize our LID as follows:
\begin{align*}
\max_{T} &|T|_{S} \quad \mathrm{s.t.,}\\
&\forall \theta' \in T, i \in [k-1], \, \mathbb{E}_{P(X_i,Y_i)}, \Big[ \phi\big(\theta', x, y\big)\Big] \leq \delta_i ,\label{eq:buffer}
\end{align*}
where $\delta_i$ is the performance requirement for the $i^{th}$ task. Importantly, this avoids the intersection step described in the previous section. Typically, replay methods try to make use of as little data as possible, meaning that expectations must be estimated with only a few samples, which can be done using the tools outlined in Subsection \ref{subsec:concentration}. 

\subsubsection{LID Biasing}\label{para:biasingLIDs}
Despite formalizing the \textit{maximal} locally invariant domain problem in Definition \ref{def:MLID}, the maximization is posed with respect to current task. Even if one computes the exact maximal LID, the domain may not span directions that are important for improving performance on subsequent tasks. Identifying such important directions has been the focus of many prior continual learning approaches. In particular, we draw from weight pruning literature, which uses weight importance measures to determine parameters that significantly contribute to the performance of a given model \cite{Mallya2018PackNet, golkar2019CLNP}. Below, we discuss how these importance methods, broken down into weight pruning or regularization, can be used to bias LID computation.

Weight pruning approaches define a discrete set of weights that are determined to be important. Formally, let $\mathcal{I}(\cdot)$ be a function associated with a weight pruning method that maps parameter indices to either 0 (if non-important) or 1 (if important) as determined by the pruning method.  We can then restrict our prior optimization formulation to only be over only the unimportant parameter subspace i.e.., leaving all important weights fixed. Where we denote members of this subspace with $\mathbb{R}^{p}_{\mathcal{I}} \subseteq \mathbb{R}^{p}$ we can reformulate the optimization problem as
\[
\max_{T \in \mathbb{R}^{p}_{\mathcal{I}}} |T|_S \quad s.t. \quad \max_{\theta' \in T} \left[ \frac{1}{N} \sum_{i=1}^N \phi(\theta', x_i, y_i) \right] \leq \delta.
\]
Intuitively, this modification only allows for updates to those weights that are unimportant for the current task, making it less likely that one violates the stated constraint (after $s.t.$ above). A critical parameter in weight pruning methods is the proportion, $\alpha \in (0, 1)$ of weights to be pruned---we explore this hyper-parameter in detail in Section~\ref{sec:experiments}.

Additionally, one might consider a softer approach to constraining the LID optimization where modification of important weights is not explicitly restricted but discouraged. For this one might use the regularization approaches discussed in the continual learning literature. Formally, we can define $\mathcal{R}(\cdot)$ to be a function that assigns a cost (a negative value) to domains that allow large updates to important parameters. For example, where per-parameter importance is captured in a vector $\beta \in \mathbb{R}^p_{+}$ and magnitude is measured the largest change allowed by $T$ per parameter, $d_T \in \mathbb{R}^p_{+}$, one could define $\mathcal{R}(T) = - \beta^\top d$. Such regularization approaches can be directly used to regularize our optimization formulation: 
\[
\max_{T \in \mathbb{R}^{p}} |T|_S  + \mathcal{R}(T)\quad s.t. \quad \max_{\theta' \in T} \left[ \frac{1}{N} \sum_{i=1}^N \phi(\theta', x_i, y_i) \right] \leq \delta.
\]

\textbf{Lookahead buffers.} As stated, the above methods rely  solely on current task data to bias the LID computation towards good solutions. While this may be effective, it still encodes assumptions that may not hold, e.g., that the important weights for this task should not be important for the next task. We can more directly bias the LID computation for good solutions if given a small amount of data (approximately) drawn from future task distributions. We refer to this method as \textit{lookahead} buffer.

The lookahead approach enables us to estimate the subspace of parameters that is important to the \textit{next} task.
Given the data in the buffer are representative of the future task, each of the biasing approaches above (weight pruning and isolation) can be leveraged, but rather than computing weight importance on the current task and discouraging modification, one can compute weight importance for the future task and reward domains that enable large magnitude modifications to \looseness=-1 these weights.

\subsubsection{Finite-Sample Guarantees for $\delta$-Safe Updates}\label{subsec:concentration}

Definition~\ref{def:provable-update} involves a constraint with an intractable integral over the distribution $P(X, Y)$, which is typically taken to be unknown. In practice, we approximate the expectation using a finite evaluation set that is held out during training (or held out within a data buffer): $\{(x^{(i)}, y^{(i)})\}_{i=1}^N \sim P(X,Y)$, with the empirical estimate denoted as
$$\hat{\phi}_N := \frac{1}{N} \sum_{i=1}^N \phi\big(\theta + \mathcal{M}(\theta, \mathcal{D}'), x^{(i)}, y^{(i)}\big).$$
Assuming finite variance of this estimator, Chebyshev’s inequality can be applied, giving: $\Pr\left( \left| \hat{\phi}_N - \mathbb{E}[\phi] \right| \geq \epsilon \right) \leq \frac{\sigma^2}{N \epsilon^2}.$
For any $\beta \in (0,1)$, solving for $\epsilon$ such that the right-hand side equals $\beta$, gives that with probability at least $1 - \beta$:
\[
\mathbb{E}_{P(X,Y)}\left[\phi\left(\theta + \mathcal{M}(\theta, \mathcal{D}'), x, y\right)\right] \leq \hat{\phi}_N + \sqrt{\frac{\sigma^2}{N\beta}}.
\]
Therefore, to certify $\delta$-safety with confidence $(1-\beta)$, it suffices that:
$\hat{\phi}_N \leq \delta - \sqrt{\frac{\sigma^2}{N\beta}}.$
If the risk metric is bounded $\phi(\cdot) \in [a,b]$, we can tighten this bound using Hoeffding's inequality to: $\hat{\phi}_N \leq \delta - \sqrt{{(b - a)^2 \log(1/\beta)}/{(2N)}}$.

\looseness=-1However, if $\phi$ is an unbounded function (e.g., cross-entropy) and the variance is unknown, strict certification becomes challenging. In such cases, we can rely on the Central Limit Theorem (CLT) to provide a probabilistic argument for safety, approximating the error of an empirical estimate as $z_{1-\beta}\frac{\hat\sigma}{\sqrt{N}}$ where $z_{1-\beta}$ is the $z$-score at $(1-\beta)$ and $\hat\sigma$ is the sample standard deviation. We note, however, while this is effective in practice for large sample sizes $N$, it remains an asymptotic approximation.

To ensure strict, provable finite-sample safety, we restrict our formal guarantees to bounded specification functions where $\phi(\cdot)\in[a,b]$. For nominally unbounded objectives, we employ a clipped version $\phi_\mathrm{clipped}=\max(L,\min(\phi,U))$ where $U$ forms the upper bound and $L$ the lower bound. This ensures the variance is strictly bounded, validating the use of Chebyshev's inequality and further tightening arguments using Hoeffding's inequality. In our experiments, we utilize accuracy (bounded in $[0,1]$), ensuring these finite-sample guarantees hold. While a theoretically valuable discussion, in practice most specification functions are naturally bounded or can be reasonably artificially bounded.

We emphasize that without such guarantees being formally stated, LID computation may be unsound. For example, without a finite-sample correction, a LID computed with safety constraints from a single held out test point could encompass a much larger parameter domain, but will ultimately merely yield a trivial safety guarantee. Moreover, the requirement for our concentration argument that sample points are i.i.d. from the joint distribution $P(X, Y)$ rules out any constraint sampling heuristics. 

\section{Practical Computations for Safe Updates}

We now present our computational framework for estimating maximal LIDs. Directly solving the optimization problem \eqref{eq:mlid} over arbitrary connected domains $T$ is computationally intractable. To obtain tractable approximations, we introduce two relaxations. First, we restrict $T$ to a parametric family of abstract domains, denoted $T(\alpha)$, where $\alpha$ encodes the domain parameters (e.g., widths of orthotopes or linear constraints of zonotopes). Second, we approximate the inner expectation by an empirical average over a finite sample. These relaxations yield the following bi-level optimization problem:

\begin{equation}
\max_{\alpha} \left|T(\alpha)\right|_{S} \, \, s.t. \, \,  \max\limits_{\theta' \in T(\alpha)} \left[ \, \frac{1}{N}\sum_{i=1}^N \phi(\theta', x_i, y_i)\right] \leq \delta.\label{eq:bilevel_opt}
\end{equation}

In the subsequent sections, we outline the propagation of our chosen abstract domains and describe our primal-dual algorithm for solving this formulation.

\subsection{Propagation of Abstract Domains}

The inner maximization in~\eqref{eq:bilevel_opt} corresponds to a \textit{parameter-space verification problem}, which seeks to upper bound the empirical loss over a set of model parameter values. This formulation is analogous to recent work on verifying robustness to parameter-space perturbations~\cite{wicker2022robust, sosnin2024certified}, but differs from standard adversarial robustness settings that consider perturbations over the input space. Specifically, we are concerned with the worst-case model outputs over all parameter vectors rather than input vectors.
For ease of exposition, we discuss the case where $T$ is the \textit{interval domain}, denoted $T(\alpha)$ with $\alpha :=[\alpha^L, \alpha^U]$, which encodes all parameter vectors $\theta \in \mathbb{R}^p $ satisfying
\begin{equation*}
\theta \in T(\alpha) \quad \iff \quad \forall i \in \{1, \ldots, p\}, \quad \alpha^L_i \leq \theta_i \leq \alpha^U_i.
\end{equation*}

This representation defines a box in parameter space and serves as an abstract domain, formally known as the orthotope domain, over which we can compute bounds on the output of a neural network using Interval Bound Propagation (IBP) \cite{gowal2018effectiveness}.

Given a neural network $f^\theta$ and an interval domain $T(\alpha)$, IBP computes lower and upper bounds on the network output at an input point $x$ for any $\theta \in T(\alpha)$. 
This propagation relies on all used activation functions being monotonic (e.g., ReLU, tanh) \cite{gowal2018effectiveness}. This assumption should not be substantially restrictive in practice, as most commonly used activation functions are monotonic. Moreover, there are some ways to relax this requirement.
Subsequently, these output bounds can then be propagated through the specification function $\phi$ to obtain a sound upper bound on the empirical specification estimate. In particular, we use the following IBP-based relaxation:
\begin{equation*}
\max_{\theta \in T(\alpha)} \frac{1}{n}\sum_{i=1}^n \phi\left(\theta, x_i, y_i\right)
\;\; \leq \;\; \frac{1}{n}\sum_{i=1}^n \phi_{\text{IBP}}\left(T\left(\alpha\right), x_i, y_i\right),
\end{equation*}
where $\phi_{\text{IBP}}(\cdot)$ denotes the IBP-derived upper bound on $\phi\left(\theta, x_i, y_i\right)$ valid for all $\theta \in T\left(\alpha\right)$.
The IBP approximation is both \textit{sound} (it provides a guaranteed upper bound) and \textit{computationally efficient}, requiring at most 4$\times$ the cost of a standard forward pass~\cite{gowal2018effectiveness}. 
Moreover, the IBP procedure is differentiable with respect to the parameters $[\alpha^L, \alpha^U]$, enabling gradient-based optimization of the outer problem.

\begin{figure*}
    \centering
    \includegraphics[width=1.0\textwidth]{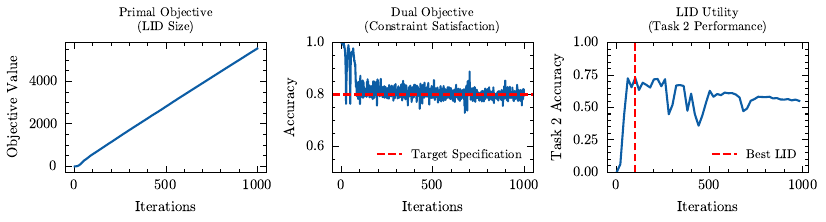}
    \caption{Analysis of primal-dual optimization for LID computation. We observe that although the primal-dual optimization continues to find larger LIDs according to the primal objective, the LIDs with the best utility occur early on in optimization.}
    \label{fig:convergence}
\end{figure*}

\subsection{Primal-Dual Optimization for Computing Maximal Locally Invariant Domains}

Interval domains offer not only a natural parameterization for the constrained problem in Equation ~\eqref{eq:bilevel_opt}, but also a differentiable and sound over-approximation of the constraint. This yields a single-level constrained optimization problem that constitutes a \textit{sound under-approximation} of the original formulation. That is, any solution identified via this relaxation defines a LID that satisfies the safety specification, although it may be strictly smaller than the solution of the original bi-level problem in Equation ~\eqref{eq:bilevel_opt}.

Although the relaxed problem still does not admit an analytic solution, it is amenable to Lagrangian-based optimization. Specifically, to maximize the size of the abstract domain subject to the IBP-based constraint, we formulate the following Lagrangian:

\begin{equation*}
L(\alpha, \lambda) = \left|T(\alpha)\right|_S + \lambda \left(\delta - \frac{1}{n}\sum_{i=1}^n \phi_{\text{IBP}}\left(T\left(\alpha\right), x_i, y_i\right)\right),
\end{equation*}

where $\lambda \geq 0$ is the Lagrange multiplier for the (bounded) safety specification constraint.

This gives rise to the saddle-point optimization $\max_{\alpha} \min_{\lambda \geq 0} L(\alpha, \lambda)$ which can be locally solved using \textit{alternating gradient descent-ascent} \cite{zhang2022near}, which performs the updates:
\begin{align}
\lambda_{t+1} &= \left[\lambda_t + \eta_d \nabla_\lambda L(\alpha_t, \lambda_t)\right]_+, \\
\alpha_{t+1} &= \alpha_t - \eta_p \nabla_{\alpha} L(\alpha_t, \lambda_{t+1}),
\end{align}
where $\eta_d$ and $\eta_p$ are step sizes for the dual and primal updates, respectively, and $[\cdot]_+$ denotes projection onto the non-negative orthant to ensure $\lambda \geq 0$\footnote{We employ the \texttt{cooper} library for Lagrangian-based optimization~\cite{gallego2025cooper}.}. We additionally apply an extra projection step to $\alpha$ that enforces that the nominal parameters $\theta$ remain inside $T(\alpha)$. The above primal-dual optimization procedure is designed to maintain feasibility with respect to the safety specification constraint throughout training. However, due to the inherent stochasticity of gradient-based optimization, the final iterate may not strictly satisfy the original constraint. To formally verify that the returned LID, denoted $T(\alpha)$, satisfies the desired specification, we compute a post-optimization certificate by computing $\max_{\theta' \in T(\alpha)} \frac{1}{n} \sum_{i=1}^n \phi(\theta', x_i, y_i)$.

\begin{figure*}[t]
    \centering
    \includegraphics{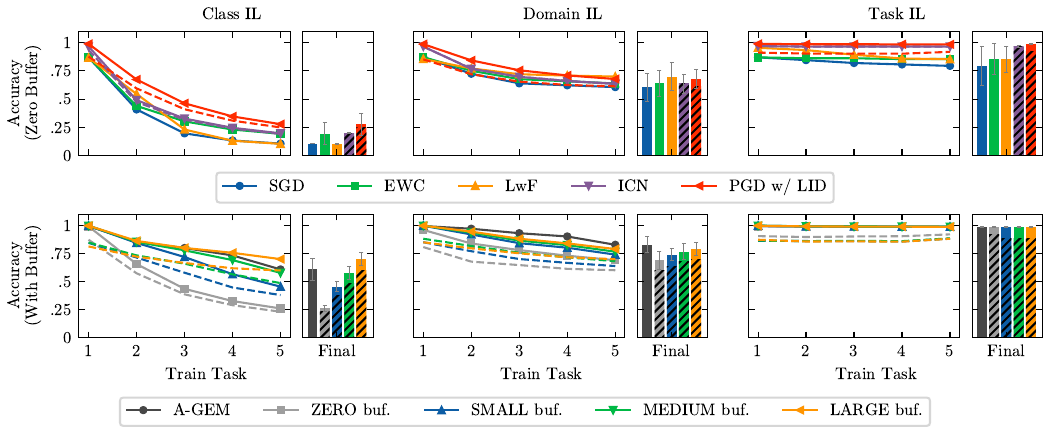}
    \caption{
    Performance on the Split-MNIST dataset under three continual learning scenarios. The top panel shows the performance of our method with zero buffer, while the bottom panel shows the performance with various buffer sizes. The dashed line and hashed bars display the certificates on performance. Error bars illustrate the 80\% confidence interval over 100 runs for the zero buffer and 15 runs for the buffer case. }
    \label{fig:split_mnist_zero_buffer}
\end{figure*}

\subsection{Choice of Specification Function} Effectively applying \textit{alternating gradient descent-ascent} algorithms and solving Equation ~\eqref{eq:bilevel_opt} imposes specific requirements on the specification function $\phi$. Most notably, any specification function must be differentiable. While a convenient choice for $\phi$ is accuracy, due to its interpretability, we cannot directly use accuracy since it is non-differentiable. Thus, our experiments rely on using soft accuracy $\phi_\mathrm{soft}$ as a differentiable surrogate:
\[
    \phi_\mathrm{soft}(\theta)=\frac{1}{N}\sum_{k=0}^N\sum_{c=1}^Cy_{k,c}\text{Softmax}(f^\theta(x_k))_c,
\]
where $y_k$ is taken to be the one-hot encoding vector of the correct class label.

Although soft accuracy, or any differentiable surrogate, serves as the specification function during maximal LID computation, the final certificate of the LID must be based on the true specification function (e.g., minimum accuracy over the LID). This distinction preserves the correctness of our approach. For further discussion we refer to Appendix \ref{app:heuristics}.

\subsection{Early Stopping and Checkpointing}
One specific improvement we propose---alongside less notable, further adaptations found in Appendix~\ref{app:ibp}---is early stopping and check-pointing based on findings in Figure~\ref{fig:convergence}.
Figure~\ref{fig:convergence} shows an example LID computation using our primal-dual optimization. We observe that the primal objective (the LID size metric) does not converge within 1000 iterations. On the other hand, by inspecting the LIDs at each step of optimization, we observe that the LIDs with the best downstream performance on Task 2 occur early on in the LID optimization. This suggests that the LID computation tends to prioritize weights that do not contribute significantly to the model's output while keeping decisive weights for the next task frozen (i.e., assigning them a very small interval).

To mitigate this effect our implementation saves the current LID periodically during the computation and terminates the LID optimization before convergence. This produces a set $\mathcal{O}$ of domain parameters each corresponding to a LID. We then issue a certificate of the minimum accuracy for each LID in $\mathcal{O}$. When training the next task, we must project the model parameters (the result of proposed updates) $\theta$ onto the set of LIDs $\mathcal{O}$. Several strategies could be used to choose the LID $o^{*}$ on which we project $\theta$. An intuitive extension of the general case above is projecting onto the closest LID using 
\[
o^* = \arg \min_{o\in\mathcal{O}} \|\Pi(\theta, o) - \theta\|_2^2,
\]
where $\Pi$ is the projection operator defined as 
\begin{align*}
\Pi(\theta, o)_i = \mathrm{Clamp}(\theta_i, \theta^L_i, \theta^U_i), \qquad o = [\theta^L, \theta^U].
\end{align*}

Another strategy is to choose the LID $o^*$ that results in the best immediate performance on our current batch. More specifically, letting $\mathcal{B} := \{ (x^{(i)}, y^{(i)} \}_{i=1}^{B}$ be a batch of data we have that
\[
o^* = \arg \min_{o\in\mathcal{O}} \mathcal{L}(\Pi(\theta, o), \mathcal{B}),
\]

where $\mathcal{L}(\theta, \mathcal{B})$ denotes the loss over the batch at $\theta$.

\paragraph{Downstream Advantages of Checkpointing}
The introduction of checkpointing forced an adaptation in our overall algorithm to allow for multiple LIDs for a single model. This can be combined with aforementioned biasing methods to compute distinctly biased LIDs that can be adapted to different types of downstream tasks. Having LIDs that are expanded along separate dimensions of the parameter space, lends a significant amount of flexibility to our algorithm that predecessors like InterContiNet did not exploit \cite{wolczyk2022continual}.

While we do not further explore the practical impact of obtaining differently biased LIDs for a single model, we consider this a considerable advantage of isolating the LID problem. LID multiplicity may be a fruitful area of study for future works.

\begin{figure*}
    \centering
    \includegraphics{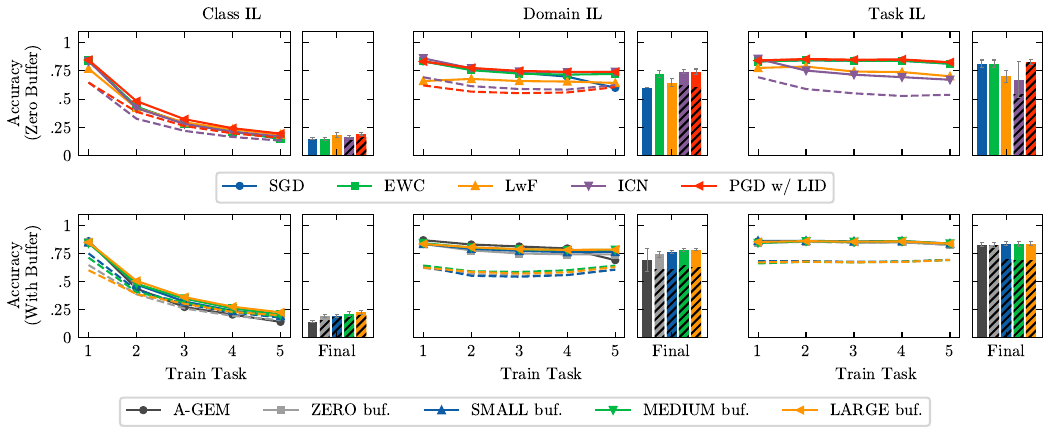}
    \caption{
    Performance on the Split-CIFAR10 \cite{CIFAR} dataset under three continual learning scenarios. The top panel shows the average accuracy over all seen tasks without replay data, while the bottom panel the average performance over seen tasks, while having varying amounts of buffer data available. The dashed line and hashed bars display the certificates returned by our method. Error bars illustrate the 80\% confidence interval over 15 runs.}
    \label{fig:split_cifar_zero_buffer}
\end{figure*}

\subsection{Algorithm Discussion}

\paragraph{Strengths} A major strength of our proposed framework is its compatibility with a wide range of prior research advances. In particular, we can leverage a large number of existing approaches in forgetting-safe continual learning and in alignment-preserving fine-tuning. Moreover, given that our method is independent of how one arrives at $\theta$ for their initial task, our method can be used, in principle with any machine learning model. This means we can accommodate arbitrarily large foundation models. Additionally, because maintaining our guarantees only requires the introduction of a projection operator, we can apply our approach to any proposed updating scheme. This is particularly advantageous given the rapid rate at which model updating algorithms advance, e.g., RLHF algorithms such as DPO, PPO, and GRPO. Finally, our approach relying on abstract interpretation is computationally efficient and is able to certify a LID, expressed as an interval domain, in $4\times$ the cost of a simple forward pass.

\paragraph{Limitations} Despite its computational efficiency and differentiability, the IBP approximation we leverage is known to exhibit increasing looseness with network depth and width~\cite{katz2017reluplex, sosnin2024certified}.
Thus, while arbitrarily large networks can be considered, parameter intervals over the early layers of the network contribute disproportionately to the overall interval bounds on the outputs \cite{wicker2020probabilistic}.
As a result, our LIDs likely exhibit larger intervals for the later layers of the network compared to the earlier layers. In sufficiently deep networks, this effect can lead to the early layers becoming effectively `frozen,' with little (or no) flexibility for future adaptation, while the final layers, subject to less over-approximation, remain relatively less constrained. 
This difficulty means that ensuring useful guarantees for large models requires careful design and hyper-parameter choices; we provide further systematic discussion of this phenomenon in Appendix~\ref{app:heuristics}.

\paragraph{Potential Future Applications} 
In this section we discuss some of the potential future applications enabled by a framework that enables \textit{a priori} safe update guarantees. Most directly, the requirement for an \textit{a priori} safe update mechanism has been put forth as a regulatory requirement for adjusting models in some critical domains. For instance, the U.S. Food and Drug Administration (FDA) outlined guidelines that enable pre-planned model updates, but only if they can be demonstrated to be safe \textit{a priori} \cite{us2024predetermined}. Furthermore, regulatory attention has also been attracted by the wide-spread use of foundation models (e.g. large language models). Regulations surrounding foundation models may require persistent safety mechanisms, especially given that models must respond dynamically to emerging vulnerabilities. While a valid response would be model de-provisioning (taking the model offline), using known-safe model updates is preferable for availability-critical systems. In the context of updating provided or open-source large language models, a potential use case of our method could be guaranteeing the preservation of learned behavior (e.g. alignment) for down-stream fine-tuners. As we have discussed, if not done carefully, fine-tuning can rapidly erode alignment properties \cite{anderljung2023frontier, qi2024evaluating}. Thus, allowing third-party users to safely tune a model while the publishing entity can ensure, \textit{a priori}, that this individualized updating procedure does not yield a model that violates originally aligned behavior. Such a guarantee may lead to robust terms-of-use. \looseness=-1

\section{Experiments}\label{sec:experiments}\label{sec:experimental}

We empirically validate our approach on standard continual learning benchmarks and provide further experimental details and ablations in Appendix~\ref{app:ablation}. All experiments were performed on compute infrastructure equipped with 2x NVIDIA L40 GPUs, using a custom Pytorch implementation. Our codebase will be publicly released upon publication.

\begin{figure*}[ht]
    \centering
    \includegraphics[width=0.95\linewidth]{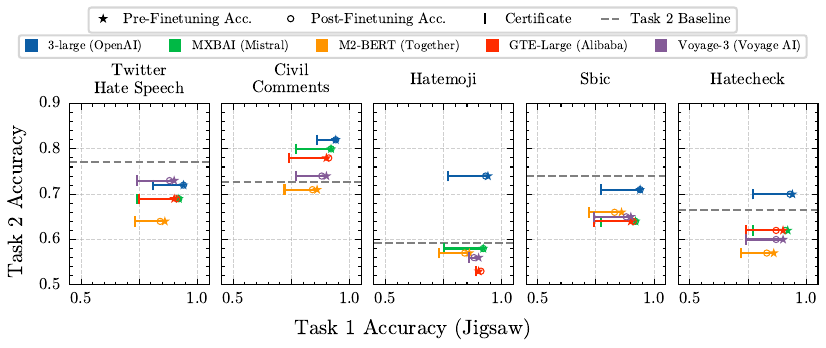}
    \caption{ Performance of fine-tuning with LIDs in the Hate-Speech Classification setting, where certificates are given with respect to the first task.
    Each color corresponds to a different LLM embedding model with the company of origin in parentheses. 
    The gray dashed line denotes the average baseline performance on the fine-tuned task when directly using the M2-BERT model..}
\label{fig:hatespeech}
\end{figure*}

\subsection{Continual Learning Experiments}\label{subsec:cl_experiments}

\textbf{Baselines} We compare our method against the following established continual learning methods: (1) EWC \cite{kirkpatrick2017overcoming}: a regularization approach that selectively weights important parameters based on Fisher information, (2) LwF \cite{li2017learning}: a distillation-based method that preserves functional behavior by using current task samples, (3) SGD: sequential training across tasks without explicit mitigation of catastrophic forgetting and (4) InterContiNet (ICN): a weight-space-constraint based approach to avoiding catastrophic forgetting \cite{wolczyk2022continual}. To benchmark our buffer extension, we observe our method compared to its standard, zero-buffer implementation and Averaged GEM (A-GEM) \cite{chaudhry2019efficientlifelonglearningagem}.
For each baseline we consider the main incremental learning protocols outlined in \cite{van2022three}:
\begin{itemize}
    \item \textbf{Class-IL:} The model learns to discriminate between an expanding set of classes. Each context introduces a new set of classes (e.g., first context contains `0' and `1', second context contains `2' and `3', etc.), and the model must classify across all previously encountered classes without task identity information at test time.
    \item \textbf{Domain-IL:} The underlying task remains constant while the input distribution shifts across contexts. The model performs binary classification (odd vs. even digits in Split-MNIST and animal vs. non-animal in Split-CIFAR), but each context presents previously unseen input domains. Task identity remains unavailable during inference. \looseness=-1
    \item \textbf{Task-IL:} Learning to classify an expanding set of classes, but only within the boundaries of each task context. Critically, task identity is provided during inference, allowing the model to constrain its prediction space appropriately.
\end{itemize}

\paragraph{Evaluations} In Figures~\ref{fig:split_mnist_zero_buffer} and \ref{fig:split_cifar_zero_buffer} we compare the performance of our approach and baselines. In the top row of each figure we plot the average accuracy/certificate on all previously seen tasks (y-axis), as we observe the current task (x-axis). In the bottom row, we show the performance of the buffer-extension to our approach using a small (1000 samples), medium (5000 samples), and large (15000 samples) buffer. Additionally, we compare our method with A-GEM with a large buffer and present results in a manner similar to the top row plots. \looseness=-1

\paragraph{Split-MNIST} We divide the standard MNIST dataset \cite{lecun1998gradient} into five sequential contexts, each containing an odd and an even number. Digit allocation across contexts was randomized for each experimental run, with reported performance metrics averaged across 100 independent runs. Figure~\ref{fig:split_mnist_zero_buffer} presents our method's performance across the three Split-MNIST scenarios. We train a feedforward neural network with 2 convolutional layers (each with 8 channels and a $5\times5$ kernel), followed by a dense layer. In Task-IL, all continual learning approaches effectively acquire new tasks while preserving performance on previous ones. However, for the more challenging Class-IL and Domain-IL settings, average accuracy deteriorates as task count increases. 
Impressively, our approach demonstrates competitive performance in the Class-IL setting, while simultaneously providing theoretical guarantees (dashed line) that bound the minimum accuracy under any future learning. In Domain-IL, our method maintains competitive average performance compared to baseline approaches while exhibiting improved retention of first-task performance. Moreover, our method observes more flexibility in its constraints, whereas ICN is highly prone to constraint collapse, as evident by the depicted tightness of the bounds.

\begin{figure*}
    \centering
    \includegraphics[width=0.95\linewidth]{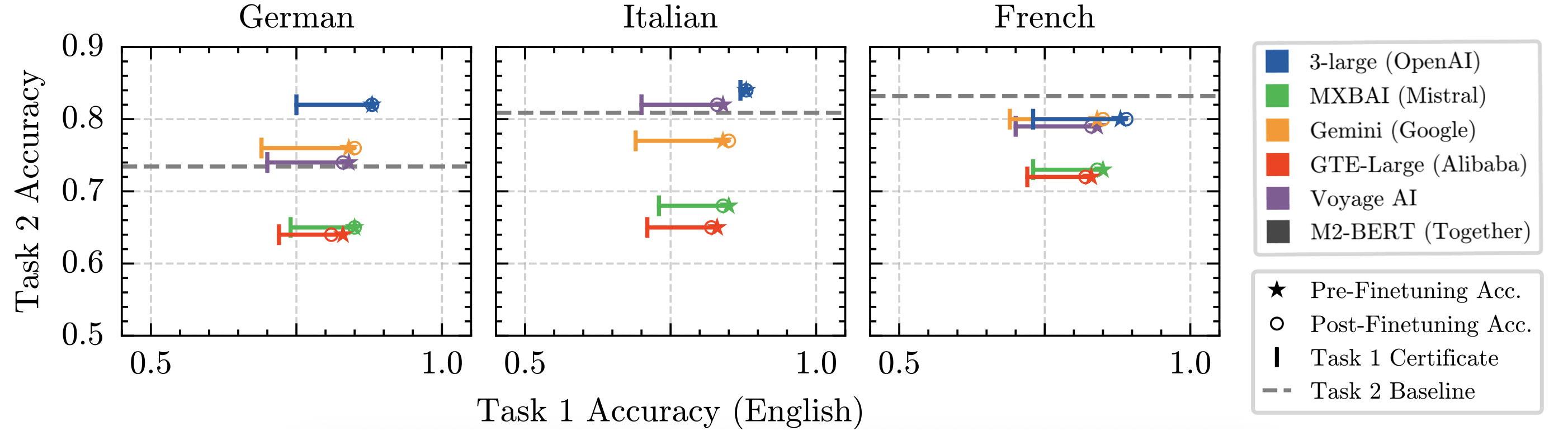}
    \caption{Performance of fine-tuning with LIDs on multilingual sentiment analysis task. We update a sentiment analysis classifier originally trained on english language reviews on multilingual reviews. We observe non-trivial guarantees on all benchmarks (vertical line) while maintaining performance on task 1 (hollow dot, x-axis) and besting baseline performance (dashed line, y-axis).}
    \label{fig:multilingual}
\end{figure*}

Additionally, using buffer data can drastically increase the performance of our method with a large buffer outperforming even A-GEM in the most difficult continual learning setting, CIL, and strictly improving performance in Domain-IL. Task-IL does not show the same trend as the setting is too simple and allows for enough model flexibility for even the zero buffer case to extract consistently high performance.

\paragraph{Split-CIFAR} We now turn to the Split-CIFAR10 dataset and consider the training and subsequent fine-tuning of a pre-trained ResNet18 model. The entire model is first trained on the initial task, after which subsequent tasks are learned by fine-tuning only the final linear layer of the network. The 10 classes are randomly allocated to 5 tasks, following the three incremental learning settings described above. In Figure~\ref{fig:split_cifar_zero_buffer}, we observe similar overall trends to Split-MNIST; specifically, our method is consistently competitive with baseline approaches, while also offering formal guarantees on forgetting performance. In Domain-IL, we observe consistently better accuracy than baselines though our worst-case guarantees are below the performance of baseline methods (which is generally to be expected). Class incremental learning remains extremely challenging for all methods---where we consider only preserving task 1 performance, our method significantly out-performs baselines. \looseness=-1

Following the findings on Split-MNIST, our method further distinguishes itself from InterContiNet on Split-CIFAR10, where ICN lacks significant adaptability to different task settings due to its sensitivity to hyperparameters and numerical instability.

\subsection{Fine-Tuning Experiments}
Previous experiments have primarily focused on multi-step updates, we now further explore our method on single-step updates via foundation model fine-tuning.
\subsubsection{Foundation Model Performance on Hate-Speech Classification}\label{subsec:llm_experiments}

We evaluate our approach on the binary task of Hate-Speech Classification (non-hate, hate labels) for the following datasets: Jigsaw \cite{jigsaw-toxic-comment-classification-challenge}, Twitter
Hate Speech \cite{davidson2017automated}, Civil Comments \cite{civil_comments}, Hatemoji \cite{hatemoji}, Sbic \cite{sbic} and Hatecheck \cite{hatecheck}. For that purpose we use the frozen embeddings of the following foundation models that at the time of writing have top performance on existing benchmarks : 3-large (OpenAI) \cite{openai2024embedding}, MXBAI (Mistral) \cite{choi2024linq}, M2-BERT (Together) \cite{together2024m2bert}, GTE-Large (Alibaba) \cite{ye2022complementary}, Voyage-3  (Voyage) \cite{voyage2024voyage3}.

For each foundation model, we proceed by first training a linear layer in an unconstrained manner on the frozen embeddings of that model on the Jigsaw dataset. For each model, we then certify approximately maximal LIDs for the linear layer assuming a macro-accuracy (also known as balanced accuracy) constraint on Jigsaw.

\begin{figure}
    \centering
    \includegraphics[width=1\linewidth]{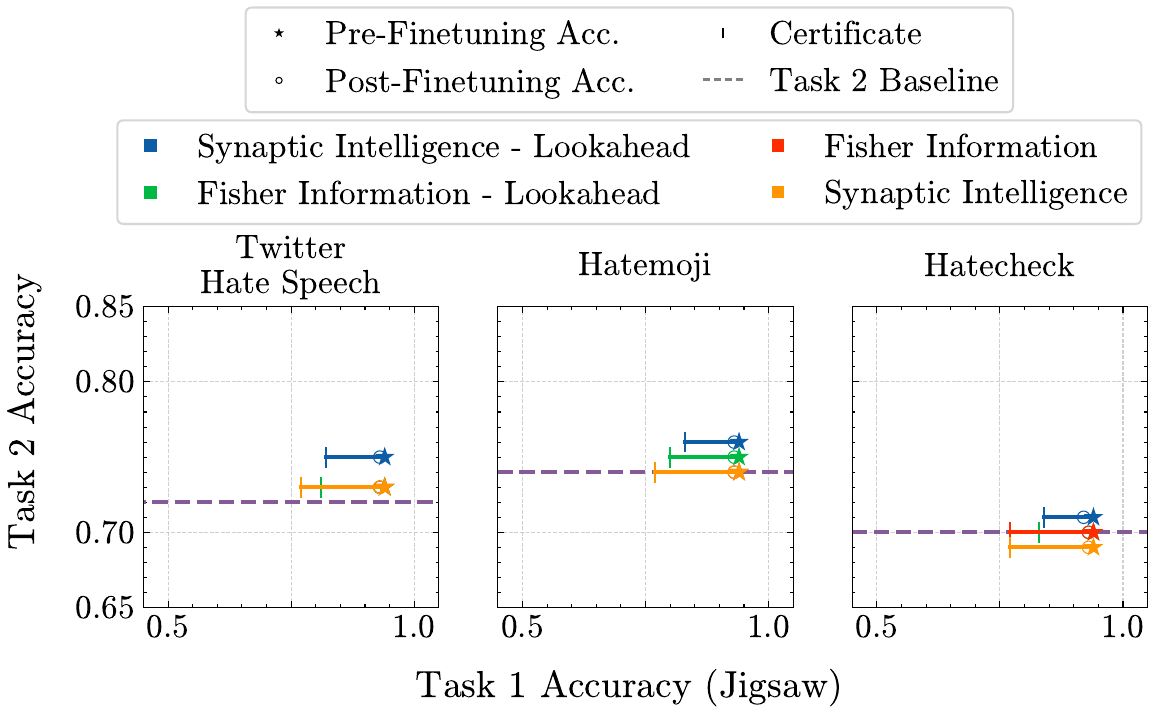}
    \caption{Performance of various biasing metrics applied to LLM head fine-tuning for hate speech classification. All data points are based on 3-large (OpenAI) as the underlying foundation model. The baseline corresponds to certified fine-tuning without any biasing.}
    \label{fig:llm_lookahead}
\end{figure}

\begin{table*}[ht]\footnotesize
    \centering
    \caption{Task 2 accuracy comparison across different pruning proportions with and without lookahead buffers. Values are reported as Mean $\pm$ Std Dev. The baseline performance for an unbiased LID is 0.20.} 
    \resizebox{\textwidth}{!}{%
    \begin{tabular}{lccccccc}
        &&&\textit{With Lookahead Buffer}  \\
        \toprule
        \textbf{Pruning Prop.} & \textbf{0.5} & \textbf{0.7} & \textbf{0.8} & \textbf{0.825} & \textbf{0.85} & \textbf{0.9} & \textbf{0.95} \\
        \midrule
        Synaptic Intelligence & $0.37 \pm 0.22$ & $0.53 \pm 0.29$ & $0.67 \pm 0.19$ & $0.73 \pm 0.18$ & $0.65 \pm 0.19$ & $0.39 \pm 0.22$ & $0.07 \pm 0.14$ \\
        Fisher Information & $0.35 \pm 0.18$ & $0.45 \pm 0.31$ & $0.64 \pm 0.26$ & $0.73 \pm 0.15$ & $0.67 \pm 0.12$ & $0.38 \pm 0.25$ & $0.05 \pm 0.12$ \\
        \midrule
        &&&\textit{No Lookahead Buffer (Standard)} \\
        \toprule
        \textbf{Pruning Prop.} & \textbf{0.5} & \textbf{0.3} & \textbf{0.2} & \textbf{0.175} & \textbf{0.15} & \textbf{0.1} & \textbf{0.05} \\
        \midrule
        Synaptic Intelligence & $0.14 \pm 0.11$ & $0.37 \pm 0.20$ & $0.28 \pm 0.18$ & $0.28 \pm 0.17$ & $0.38 \pm 0.18$ & $0.38 \pm 0.26$ & $0.39 \pm 0.24$ \\
        Fisher Information & $0.14 \pm 0.10$ & $0.35 \pm 0.21$ & $0.31 \pm 0.21$ & $0.31 \pm 0.22$ & $0.36 \pm 0.23$ & $0.39 \pm 0.21$ & $0.39 \pm 0.25$ \\
        \bottomrule
    \end{tabular}%
    }
    \label{tab:lookahead_finding}
\end{table*}

We then evaluate our approach by updating the Jigsaw-trained model on each of the five downstream datasets. The results are summarized in Figure~\ref{fig:hatespeech}, where each subplot corresponds to a different target dataset and visualizes the performance trade-offs induced by LID-constrained updates. Along the x-axis, we show performance on the original task (Jigsaw), while the y-axis reflects performance on the new task. In particular, the x-axes show the initial performance on Jigsaw (denoted with a star), the post-finetuning performance with the LID (denoted with an open dot), and the formal lower-bound from the LID on Jigsaw macro accuracy (denoted with a bar). Performance on the new task is shown along the y-axis, with the gray dashed line indicating the baseline performance of M2-BERT with EWC.  Across all model-dataset pairs, we observe that LID-constrained updates may slightly degrade downstream accuracy, though in some cases performance is on par with or exceeding the unconstrained M2-BERT baseline is observed. Simultaneously, our method certifies non-trivial lower bounds on the original task’s macro accuracy, often guaranteeing that substantial accuracy is retained even under worst-case perturbations within the safe domain.

\subsubsection{Foundation Model Performance on Multilingual Sentiment Analysis}\label{subsec:multilingual_llm_experiments}

Following the same procedure as above, we additionally evaluate our approach on the multilingual classification task of sentiment analysis (negative, neutral, positive) for the following multilingual dataset : XLM-T \cite{xlmt} which contains 8 splits corresponding to 8 different languages. 

We provide a subset of the results in Figure~\ref{fig:multilingual}. The dotted gray line represents the baseline performance of Voyage-3 using EWC \cite{kirkpatrick2017overcoming}. When looking at the performances on Gemini, Voyage-AI and 3-large, we observe that our method achieves performances comparable to the EWC baseline and even occasionally outperforms it. Additionally, our method certifies non-trivial bounds on the original task macro-accuracy. These results provide the first empirical evidence that formal forgetting/alignment-preserving guarantees can be achieved in real-world LLM-based NLP pipelines, without sacrificing utility on the new task. The certified safety margins, though conservative, are large enough to ensure that updates remain practically useful.

\subsection{LID Biasing Experiments}
In this section we explore ablations on our proposed LID biasing approaches by considering the Split-MNIST CIL setting. We keep the convolutional architecture described in prior experiments and consider a non-biased LID computation as a baseline. We explore biasing the optimization using the Fisher information matrix \cite{aich2021elasticweightconsolidationewc} and synaptic intelligence \cite{zenke2017synaptic_intelligence} to obtain weight importance metrics by training our model on a batch with either lookahead data or the original task data. All experiments in this section consider pruning rather than regularization. \looseness=-1

In Table \ref{tab:lookahead_finding}, we show the performance on a downstream task across different pruning percentages. We re-run our results using 15 different random seeds and plot the standard deviation as our error bar. For both weight selection using Fisher information and synaptic intelligence, we observe significant gains in the average case, but with large variance. On the other hand, with a lookahead buffer, we observe that biasing can aid in downstream performance across a relatively broad range of pruning proportions. Performance seems to peak at 82.5\% and 5\% of frozen parameters for the lookahead and naive method, respectively. \looseness=-1

\begin{figure}[H]
    \centering
    \includegraphics[width=0.8\linewidth]{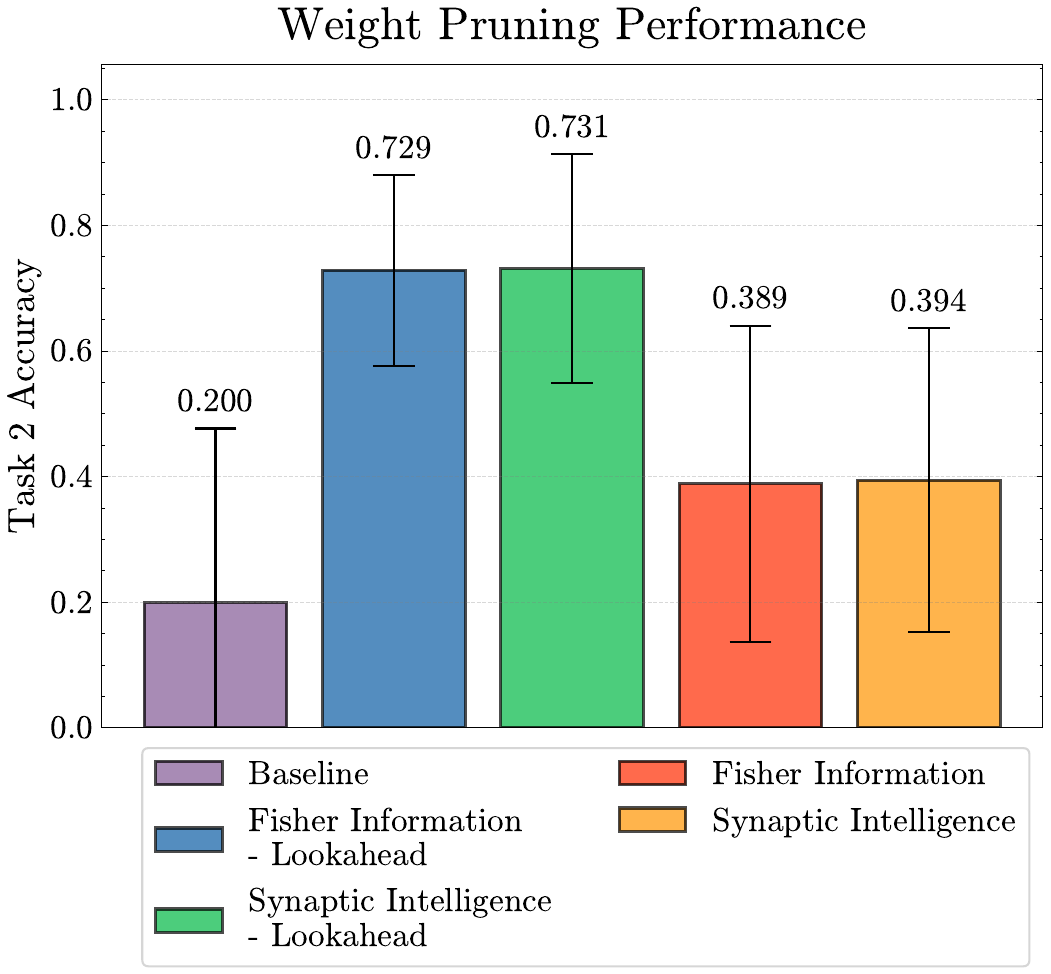}
    \caption{Performance of all of our biasing approaches for the best performing pruning percentages based on Table \ref{tab:lookahead_finding}. Pruning percentages are 82.5\% and 5\% for lookahead and no lookahead pruning, respectively. All methods are run across 15 seeds, displaying the mean performance with error bars showing the standard deviation.}
    \label{fig:lookahead_performance}
\end{figure}

Figure \ref{fig:lookahead_performance} further highlights the significance of the performance improvements for single-step continual learning scenarios (i.e. limited to two tasks) when using LID biasing. In particular, lookahead biasing unlocks a significant amount of performance by drastically simplifying the LID optimization problem. However, even the smaller amount of pruning that occurs in the non-lookahead optimization yields a simplification that is often sufficient to extract additional performance.

We additionally explore these ablations for fine-tuning LLMs. Figure \ref{fig:llm_lookahead} shows that lookahead biasing performs better than or equal to an unbiased baseline in all of the explored settings, whether it be hate speech detection or multilingual sentiment analysis. Our observations from the MNIST setting remain valid in LLM fine-tuning with LID biasing playing a relatively minor role when lookahead buffer data is not used. Interestingly, we find that synaptic intelligence with a lookahead buffer allows noticeable improvements in both formal guarantees and downstream performance.   Further experimental details can be found in Table~\ref{tab:pruning_hyperparameters} in the appendix.

\section{Conclusion}

In this work, we introduce a novel framework for computing provably safe model updates via maximal locally invariant domains (LIDs), offering formal guarantees that behavioral properties such as alignment and task performance are preserved. By leveraging abstract interpretation and a primal-dual optimization strategy, our approach enables efficient certification of model updates in both continual learning and foundation model fine-tuning scenarios. Across diverse empirical settings, we demonstrated that our method not only provides formal safety guarantees for these updates, but also matches or outperforms existing baselines in practice.

We then extend this approach by directing the optimization of locally invariant domains using biasing inspired by classical continual learning approaches. In particular, we find that biasing drastically reduces the complexity of the LID computation, yielding more effective parameter subspaces. We further show that this not only holds theoretically but also experimentally.

\section*{Acknowledgments} \noindent We would like to thank Hassen Aissa for his feedback and discussion during the initial development of this project.

\bibliographystyle{IEEEtran}
\bibliography{IEEEfull,aaai2026}

\clearpage
\newpage 

\appendix

\subsection{Additional Related Works}\label{app:related-works}

In this section, we expand upon the related works provided in the main text to discuss provable guarantees that are related to this work. Firstly, we highlight that guarantees we leverage, in the interval case, build on the algorithm proposed in \cite{rump1999fast}. This algorithm found use in neural network architectures with Bayesian neural networks \cite{wicker2020probabilistic, wicker2023adversarial} and has subsequently been used in explainability \cite{wicker2022robust}, control \cite{wicker2021certification}, quantization robustness \cite{dangcertifiably}, and in the initial work on continual learning \cite{wolczyk2022continual}. In the initial work, \cite{wolczyk2022continual}, the authors focus particularly on continual learning application and spend considerable time discussing the problem of intersecting safe regions and its complexity. In this paper, on the other hand, we focus primarily on a principled approach to computing a locally invariant domain. Thus, our problem formulation is slightly different and our perspective on the problem leads us to explore some practically important aspects not covered in \cite{wolczyk2022continual} including the use of buffer data and providing analysis of finite-sample guarantees for our method. Within our main text we reference the numerical instability of the approach given by \cite{wolczyk2022continual}, a symptom of which can be observed in their appendix where they use a learning rate hyperparameter that ranges in magnitude from $1e3$ to $1e-20$. This is due to the fact that interval bound propagation can be very loose for even moderate networks, thus as the problem scales to medium networks (CNNs even), the approach of \cite{wolczyk2022continual} will need to use vanishingly small learning rates. Similar to ICN, the approach provided in \cite{krukowski2025hint} attempts to improve the computation of LIDs using hypernetworks. We do not compare with this method as the proposal is not strictly sound due to the lack of hard constraints when updating hypernetwork weights.

\subsection{Interval Bound Propagation}\label{app:ibp}

This section details the interval bound propagation (IBP) procedure we use to compute a sound over-approximation of the specification function required for our LID computation. For clarity, we focus on feed-forward networks with monotonic activations, though the approach generalizes straightforwardly to other architectures such as convolutional networks.

\textbf{Feedforward Neural Networks.} We consider a neural network model $f^\theta: \mathbb{R}^{n_0} \rightarrow \mathbb{R}^{n_K}$ with parameters $\theta = \{(W_i, b_i)\}_{i=1}^K$ composed of $K$ layers. Each layer performs the following operations: 
\begin{align}
\hat{z}_k &= W_k z_{k-1} + b_k,  \qquad z_k = \sigma(\hat{z}_k)
\end{align}
where $z_0 := x$, $f^\theta(x) := \hat{z}_K$, and $\sigma$ is the activation function, which we assume to be monotonic (e.g., ReLU).

\textbf{Interval Arithmetic.} We now define interval representations for matrices, and their corresponding arithmetic operations. We represent each matrix $A$ by its element-wise interval bounds $[A^L, A^U]$, where $A^L$ and $A^U$ denote the lower and upper bounds respectively. The key operations required for neural network forward passes are interval matrix addition and multiplication.

Given interval matrices $[A^L, A^U]$ and $[B^L, B^U]$, we define:
\begin{itemize}[leftmargin=*]
    \item \textit{Interval Matrix Addition:} $[A^L, A^U] \oplus [B^L, B^U] = [A^L + B^L, A^U + B^U]$
    \item \textit{Interval Matrix Multiplication:} Using center matrices $A_\mu = {(A^U + A^L)}/{2}$, $B_\mu = {(B^U + B^L)}/{2}$ and radius matrices $A_r = {(A^U - A^L)}/{2}$, $B_r = {(B^U - B^L)}/{2}$, Rump's algorithm gives $[A^L, A^U] \otimes [B^L, B^U] = [C^L, C^U]$ where
\begin{align*}
C^L &= A_\mu B_\mu - |A_\mu| B_r - A_r |B_\mu| - A_r B_r \\
C^U &= A_\mu B_\mu + |A_\mu| B_r + A_r |B_\mu| + A_r B_r
\end{align*}
\end{itemize}
These operations ensure $A' \oplus B' \in [C^L, C^U]$ and $A' \otimes B' \in [C^L, C^U]$ for all $A' \in [A^L, A^U]$, $B' \in [B^L, B^U]$.
Both operations incur modest computational costs compared with their non-interval counterparts (2$\times$ for addition and 4$\times$ for multiplication).

\textbf{Bounding the Forward Pass.} Using the interval operations defined above, we can propagate bounds through the neural network. For any input $x$ and parameters $W_k \in [W_k^L, W_k^U]$, $b_k \in [b_k^L, b_k^U]$ for $k = 1, \ldots, K$, we compute interval bounds: 
\begin{align*}
[\hat{z}_k^L, \hat{z}_k^U] &= [W_k^L, W_k^U] \otimes [z_{k-1}^L, z_{k-1}^U] \oplus [b_k^L, b_k^U] \\
[z_k^L, z_k^U] &= [\sigma(\hat{z}_k^L), \sigma(\hat{z}_k^U)]
\end{align*} 
This procedure guarantees that $f^\theta(x) \in [\hat{z}_K^L, \hat{z}_K^U]$ for all valid parameter combinations within their respective intervals.

For the monotonic activation function $\sigma$, we can apply the function element-wise to both the lower and upper bounds of the input interval to obtain a valid output interval, preserving the soundness of the over-approximation throughout the forward pass.

\textbf{Bounding the Specification.} The procedure for propagating intervals through the specification function depends on the exact choice of specification function. We assume we have the interval bounds on the output logits of the network, $[\hat{z}_K^L, \hat{z}_K^U]$, computed using the procedure above. Let $y$ denote the true class index.

Standard performance metrics, such as model accuracy, are monotonically increasing in the logit of the correct class, and monotonically decreasing in the logits of all other classes. Thus, the `worst-case' output logits within the interval $[\hat{z}_K^L, \hat{z}_K^U]$ are given by
\begin{align}
    \hat{z}_K^{\text{worst}} = \hat{z}_K^{L} \cdot \mathbf{e}_{y} + \hat{z}_K^{U} \cdot (1 - \mathbf{e}_{y}),
\end{align}
where $\mathbf{e}_y \in \mathbb{R}^C$ is the one-hot vector corresponding to the true class $y$.

Thus, propagating interval bounds through common performance metrics corresponds to evaluating this worst-case output:
\begin{itemize}
    \item \textit{Cross-Entropy Loss:}
    \begin{align*}
    \phi_{\text{ce}}(\theta, x, y) &= -\log\left(\operatorname{softmax}(f^\theta(x))_y\right)  \\ 
     & \leq \phi_{\text{ce}}^{\text{IBP}} = 
    -\log\left(\operatorname{softmax}(\hat{z}^{\text{worst}}_K)_y\right),
    \end{align*}
    where the inequality reflects that $\phi_{\text{acc}}^{\text{IBP}}$ is a certified upper bound on the true loss for any $\theta$ in our interval domain.
    \item \textit{(Negative) Accuracy:} 
    \begin{align*}
    \phi_{\text{acc}}(\theta, x, y) &= -\mathbb{I}\left(\arg\max_i \{f^\theta(x)_i\} = y\right)\\
    &\leq \phi_{\text{acc}}^{\text{IBP}} = -\mathbb{I}\left(\arg\max_i \{\hat{z}^{\text{worst}}_{K, i}\} = y\right),
    \end{align*}
    where the inequality reflects that $\phi_{\text{acc}}^{\text{IBP}}$ represents a lower bound on the accuracy.
    \item \textit{(Negative) Soft Accuracy:} \begin{align*}
    \phi_{\text{soft acc}}(\theta, x, y) &= -\operatorname{softmax}(f^\theta(x))_y \\
    &\leq \phi_{\text{soft acc}}^{\text{IBP}} = -\operatorname{softmax}(\hat{z}^{\text{worst}}_K)_y,
    \end{align*} 
    which represents a certified lower bound on the soft accuracy of the model. We include the soft accuracy as a differentiable proxy of a hard accuracy specification in the primal-dual computation of LIDs.
\end{itemize}
Since our definition of LID requires $\phi$ to be bounded from above, we define the specification for accuracy and soft accuracy to be the negation of the nominal value.

\subsection{Heuristics for Effective LID Computation}\label{app:heuristics}

\textbf{De-bias regularisation}
One observation from our first experiments in the CIL setting was that the model trained on the first task is heavily biased towards the first classes (i.e. the network will result in a very confident prediction of a class from task 1 when confronted with data points from task 2). This means that there might be an important distance in parameter space that needs to be travelled before the model can start predicting other output classes, making future learning challenging within our current local invariant domain. Thus, we added a regularisation term that passes noise through the model and penalises over-confidence on any particular output. Here is a formal description of our method: Let $F$ be a noisy distribution over the input space $\mathcal{X} \in \mathbb{R}^n$. $F$ can be taken to be gaussian, uniform or anything else. We show below which distributions worked well in practice. We denote our machine learning model as parameterized functions $f^\theta : \mathbb{R}^n \to \mathcal{Y}$ with parameters $\theta \in \mathbb{R}^p$. In practice this function can be further expressed as follow: 
\[
f^\theta(x) = \arg\max_{i} h^\theta(x)
\]
where $h^\theta : \mathcal{X} \to \mathcal{\mathbb{R}^C}$ maps the input to their corresponding logits in $\mathbb{R}^C$ with $C$ being the number of classes we consider. The de-bias regularisation term $U(\theta)$ is expressed as follow : 
\begin{align*}
    U(\theta) = \frac{1}{m}\sum_{k=0}^{m} \|\mathrm{s}(h^\theta(X_k)) - \bar{\mathrm{s}}(h^\theta(X_k))\|_2 \\ X_1, ..., X_m \overset{\text{i.i.d.}}{\sim} F
\end{align*} 

where $\mathrm{s}$ is the softmax function and $\bar{\mathrm{s}}$ is the coordinate-wise mean  softmax function. This encourages the predicted probabilities of each classes to stay close to each other for noisy samples. 
During gradient descent the de-bias term $U(\theta)$ is added to the overall loss multiplied by a regularisation factor $\lambda$. 

Choosing the right distribution $F$ is decisive in ensuring that the neural network stays unbiased when confronted with new samples. Note that $F$ should be sufficiently distinct from the task distribution $\mathcal{D}_t$, otherwise learning task $t$ may effectively be compromised. In addition, $F$ should stay general enough to represent all potential inputs we might be confronted with in the future. 
In our experiments we choose $F$ to be uniform over the minimum and maximum value coordinate wise we encounter in our dataset $\mathcal{D} := \{ (x^{(i)}, y^{(i)} \}_{i=1}^{N}$. 

\begin{table*}[ht!]
\centering
\caption{Experimental Hyperparameters for Split-MNIST and Split-CIFAR.}
\label{tab:hyperparameters}
\begin{tabular}{@{}llll@{}}
\toprule
\textbf{Category} & \textbf{Hyperparam} & Split-MNIST & Split-CIFAR \\
&  & \textit{(TIL/DIL/CIL)} & \textit{(TIL/DIL/CIL)} \\
\midrule
\textit{Learning} & Learning Rate & 0.001 & 0.02 \\
& Batch Size & 64 & 128\\
& Number of Epochs & 5 & 5 \\
& $\ell_2$ Regularisation & 0.01 & 0.01\\
& De-bias Regularisation & 0.01 & 0.01\\
& Projection strategy & sample\_largest\_closest & best loss \\
\addlinespace
\textit{LID Optimization} & Primal Learning Rate ($\eta_p$) & 0.33 & 0.33\\
& Dual Learning Rate ($\eta_d$) & 0.01 & 0.01\\
& Batch Size & 400 & 400 \\
& Number of Iterations & 200 & 200\\
& LID Saving Period & 20 & 2 \\
\addlinespace
\textit{Baselines} & LwF $\lambda$ & 0.001/0.05/0.05 & 0.1 \\
& EWC $\lambda$ & 1e6 & 1e6 \\
\textit{InterContiNet} & Learning Rate & 0.01 & 0.01 \\
& Batch Size & 1e6 & 128 \\
& Epochs & 15/5/5 & 30 \\
& Radii lr & 100/1000/1 & 1/1/0.1 \\
\bottomrule
\end{tabular}
\end{table*}

\begin{table*}[ht!]
\centering
\caption{Experimental Hyperparameters for LLM Experiments}
\label{tab:hyperparameters_llm}
\begin{tabular}{@{}lllll@{}}
\toprule
\textbf{Category} & \textbf{Hyperparam} & Default & M2-BERT & MXBAI\\
\midrule
\textit{Learning} & Learning Rate & 0.0005 & 0.0005 & 0.0005 \\
& Batch Size & 64 & 64 & 64\\
& Number of Epochs & 1 & 1 & 1 \\
& $\ell_2$ Regularisation & 0 & 0.01 & 0.01\\
& De-bias Regularisation & 0 & 0 & 0\\
& Projection strategy & best loss & best loss& best loss \\
\addlinespace
\textit{LID Optimization} & Primal Learning Rate ($\eta_p$) & 0.1 & 0.03 & 0.03\\
& Dual Learning Rate ($\eta_d$) & 0.1 & 0.001 & 0.001\\
& Batch Size & 1024 & 1024 & 1024 \\
& Number of Iterations & 700 & 700 & 700 \\
& LID Saving Period & 20 & 20 & 20 \\
\bottomrule
\end{tabular}
\end{table*}

\subsection{Experimental Details and Additional Ablation Studies}\label{app:ablation}
In this chapter we explore the experimental setups further. All of the three main experiments follow similar testing and evaluation pipelines:

\textbf{Split MNIST.} As mentioned in Section \ref{subsec:cl_experiments} we use a two layer CNN follow by a single fully connected classification head. We first train this neural network on an initial task---task details are depending on the specific continual learning setup---then compute a LID for the given, trained model. Subsequently, we use projected gradient descent to train the model on each of the following tasks within the constraints provided by the LID. In-between tasks, we continue to compute LIDs for the most recently learned task, taking intersections between the existing LID and the newly computed LID to produce guarantees not only for the most recent task, but for all previous tasks. This procedure results in Algorithm \ref{alg:zerobuffer}:

\begin{algorithm}[ht]
    \caption{Certified Continual Learning with Locally Invariant Domains}\label{alg:zerobuffer}
\textbf{Input:} $f$ - NN model, $\theta'$ - initial parameter, $E$ - \# of epochs, $\phi$ - specification function, $\alpha$ - learning rate, $\delta$ - minimum accuracy.\\
\noindent \textbf{Output:} $\theta$ - Param. with acceptable error, $t$ - number of tasks completed  \\
\vspace*{-0.35cm}
    \begin{algorithmic}[1]
        \State $\mathcal{D}_1 = \{(x^{(i)}, y^{(i)})\}_{i=1}^{k} \sim P_1(X,Y)$
        \State $\theta_{1} \gets \texttt{SGD}(\theta', \mathcal{D}_1, E, \alpha)$
        \State $[\theta^{L}_1, \theta^{U}_1] \gets \texttt{LID}(\theta, \mathcal{D}_1, \phi, \delta)$
        \For{$j \in {2, ..., t}$}
            \State $\mathcal{D}_j = \{(x^{(i)}, y^{(i)})\}_{i=1}^{k} \sim P_j(X,Y)$
            \State $\theta_{j} \gets \texttt{PGD}(\theta_{j-1}, \mathcal{D}_j, E, \alpha, [\theta^{L}_{j-1}, \theta^{U}_{j-1}])$
            \State $[\theta^{L}_{j}, \theta^{U}_{j}] \gets \texttt{LID}(\theta, \mathcal{D}_j, \phi, \delta)$
            \State $[\theta^{L}_{j}, \theta^{U}_{j}] \gets [\theta^{L}_{j}, \theta^{U}_{j}] \cap [\theta^{L}_{j-1}, \theta^{U}_{j-1}] $
            \If{ $[\theta^{L}_1, \theta^{U}_1] == \emptyset$ }
                \State \textbf{return} $\theta_j$
            \EndIf
        \EndFor
        \State \textbf{return} $\theta_t$
    \end{algorithmic}
\end{algorithm}

To split MNIST into separate tasks we randomly separate the ten digits from MNIST into five tasks of two digits each, while ensuring that each of the pairs is a combination of an odd and an even number to make sure that the tasks are well defined for domain IL.

\textbf{Split CIFAR.} For split CIFAR we also follow Algorithm \ref{alg:zerobuffer} with the only difference being the task creation. Here, we hand pick well defined tasks to create the following splits: \textit{cat} vs \textit{truck}, \textit{frog} vs \textit{ship}, \textit{horse} vs \textit{automobile}, and \textit{dog} vs \textit{airplane}. To make the tasks compatible with domain incremental learning, note that each of the tasks uses an animal and a mode of transportation - two separate domains. Another slight difference for CIFAR is that we used a ResNet18 model and then train a MLP classifier on the embeddings obtained from the ResNet. We restrict our LID computations and continual learning to the MLP only as opposed to the entire model pipeline as is the case for the MNIST setup.

\subsection{Foundation Model Hyperparameters}

We present the hyperparameters used in Table \ref{tab:hyperparameters_llm}.

\subsection{Buffer-based Learning Scheme}
The algorithm for the buffer-based case is reasonably similar to the procedure described in Algorithm \ref{alg:zerobuffer}. The main difference is that instead of exiting after achieving the maximum accuracy within a given LID, we check if the achieved performance exceeds a certain target performance. If it does, we proceed to the next task, if it doesn't, we recompute the LID using samples stored in a buffer. When updating, we draw $k = 200$ samples of each task from the buffer and use the combined data to compute an updated LID. This updated LID, now centered at the most recent set of model parameters, $\theta$, ought to allow for further increases in performance on the current task, while still maintaining a safe-area with guarantees. This results in Algorithm \ref{alg:buffer}.

The hyperparameters for the buffer algorithm are largely the same as those shown in Table \ref{tab:hyperparameters} with the only addition being the aforementioned $k$, the target accuracy, and the maximum buffer calls per task shown in Table \ref{tab:buffer-hyperparameters}

\begin{table*}[ht!]
\centering
\caption{Experimental Hyperparameters for Split-MNIST and Split-CIFAR for our buffer-based algorithm.}
\label{tab:buffer-hyperparameters}
\begin{tabular}{@{}llll@{}}
\toprule
\textbf{Category} & \textbf{Hyperparam} & Split-MNIST & Split-CIFAR \\
&  & \textit{(TIL/DIL/CIL)} & \textit{(TIL/DIL/CIL)} \\
\midrule
\textit{LID Optimization} & Target Accuracy & 0.65/0.65/0.65 & 0.65/0.75/0.55 \\
 & Max Buffer Calls & 1/3/7 & 1/3/7 \\
\addlinespace
\textit{Baselines} & A-GEM memory size & 3750 & 3750 \\
\bottomrule
\end{tabular}
\end{table*}

\begin{algorithm}[ht]
    \caption{Certified Continual Learning with Locally Invariant Domains and Buffers}\label{alg:buffer}
\textbf{Input:} $f$ - NN model, $\theta'$ - initial parameter, $E$ - \# of epochs, $\phi$ - specification function, $\alpha$ - learning rate, $\delta$ - minimum accuracy, $target\_acc$ - target accuracy, $M$ - buffer, $b$ - buffer size, $acc$ - accuracy metric, $max\_calls$ - maximum number of buffer calls per task.\\
\noindent \textbf{Output:} $\theta$ - Param. with acceptable error, $t$ - number of tasks completed  \\
\vspace*{-0.35cm}
    \begin{algorithmic}[1]
        \State $\mathcal{D}_1 = \{(x^{(i)}, y^{(i)})\}_{i=1}^{k} \sim P_1(X,Y)$
        \State $\theta_{1} \gets \texttt{SGD}(\theta', \mathcal{D}_1, E, \alpha)$
        \State $[\theta^{L}_1, \theta^{U}_1] \gets \texttt{LID}(\theta, \mathcal{D}_1, \phi, \delta)$
        \State $\mathcal{D}_1^* = \{(x^{(i)}, y^{(i)})\}_{i=1}^{b_i} \sim P_1(X, Y)$
        \State $M_1 \gets D_1$
        \For{$j \in {2, ..., t}$}
            \State $\mathcal{D}_j = \{(x^{(i)}, y^{(i)})\}_{i=1}^{k} \sim P_j(X,Y)$
            \State $\theta_{j} \gets \texttt{PGD}(\theta_{j-1}, \mathcal{D}_j, E, \alpha, [\theta^{L}_{j-1}, \theta^{U}_{j-1}])$
            \State $[\theta^{L}_{j}, \theta^{U}_{j}] \gets \texttt{LID}(\theta, \mathcal{D}_j, \phi, \delta)$
            \State $[\theta^{L}_{j}, \theta^{U}_{j}] \gets [\theta^{L}_{j}, \theta^{U}_{j}] \cap [\theta^{L}_{j-1}, \theta^{U}_{j-1}] $
            \State $curr\_acc = acc(\theta, D_j)$
            \While{$n<{max\_calls } \And { not\ empty(} M_{i:j-1} {) } \And { target\_acc} > {curr\_acc}$}
                \State $\mathcal{D}_{1:j-1} = \{(x^{(i)}, y^{(i)})\}_{i=1}^{k} \sim M_{1:j-1}$
                \State $[\theta^{L}_{j-1}, \theta^{U}_{j-1}] \gets \texttt{LID}(\mathcal{D}_{1:j-1}, \phi, \delta)$
                \State $\theta_{j} \gets \texttt{PGD}(\theta_{j}, \mathcal{D}_j, E, \alpha, [\theta^{L}_{j-1}, \theta^{U}_{j-1}])$
                \State $curr\_acc = acc(\theta, D_j)$
            \EndWhile
            \State $D_j^\ast = \{(x^{(i)}, y^{(i)})\}_{i=1}^{b_i} \sim P_j(X, Y)$
            \State $M_j \gets D_j^\ast$
            \If{ $[\theta^{L}_1, \theta^{U}_1] == \emptyset$ }
                \State \textbf{return} $\theta_j$
            \EndIf
        \EndFor
        \State \textbf{return} $\theta_t$
    \end{algorithmic}
\end{algorithm}

\begin{table*}[ht!]
\centering
\caption{Experimental hyperparameters for weight pruning experiments. This acts as an extension to LLM and Split-MNIST experiments, hence the baseline uses hyperparameters specified in Tables \ref{tab:hyperparameters} and \ref{tab:hyperparameters_llm} unless mentioned otherwise.}
\label{tab:pruning_hyperparameters}
\begin{tabular}{@{}llll@{}}
\toprule
\textbf{Category} & \textbf{Hyperparam} & Split-MNIST & LLM \\
\midrule
\textit{LID Optimization} & Number of Iterations & 300 & 300 \\
 & LID Saving Period & 100 & 100 \\
 & Freezing Proportion - Lookahead & 0.825 & 0.85 \\
 & Freezing Proportion & 0.05 & 0.05 \\
\bottomrule
\end{tabular}
\end{table*}

\subsection{Empirical Computational Analysis}

To discuss the introduced computational overhead of providing safe updates we train each method found in Table \ref{tab:results} on two separate tasks of Split MNIST with an interleaved safe space computation for our method and ICN. Across 50 runs we report the mean time taken and one standard deviation of variance in seconds. It becomes apparent that our method does introduce some overhead compared to methods that do not compute safe spaces. While there is some additional delta due to the projection step during projected gradient descent, most of the overhead comes from the one-off step of the LID computation.

\begin{table}[ht!]
\caption{Empirical analysis of computational cost of different continual learning methods in seconds and one standard deviation.}
    \label{tab:results}
    \centering
    \begin{tabular}{lc}
        \toprule
        \textbf{Method} & \textbf{Value [s]} \\
        \midrule
        Ours & $43.18 \pm 2.19$ \\
        of which LID & $12.92 \pm 0.79$ \\
        \midrule
        ICN & $70.09 \pm 2.88$ \\
        of which LID & $51.81 \pm 2.13$ \\
        \midrule
        SGD & $16.08 \pm 0.82$ \\
        EWC & $15.17 \pm 1.49$ \\
        A-GEM & $20.29 \pm 0.89$ \\
        LwF & $14.96 \pm 0.80$ \\
        \bottomrule
    \end{tabular}
\end{table}

\subsection{Methodology Formalisms} \label{subsec:method_notation}

In Table \ref{tab:notation} we present a collection of notation used throughout Section \ref{sec:methods}.

\begin{table}[h]
    \centering
    \caption{Summary of Notation used in Section \ref{sec:methods}.}
    \label{tab:notation}
    \begin{tabular}{l p{0.75\linewidth}}
        \toprule
        \textbf{Symbol} & \textbf{Description} \\
        \midrule
        $\theta \in \mathbb{R}^p$ & The model parameter vector. \\
        \midrule
        $f^\theta$ & The parameterized machine learning model mapping inputs to labels. \\
        \midrule
        $\mathcal{D}$ & A dataset $\{(x^{(i)}, y^{(i)})\}_{i=1}^N$. \\
        \midrule
        $\mathcal{M}$ & An update mechanism (e.g., SGD) proposing a parameter update. \\
        \midrule
        $\phi(\theta, x, y)$ & The specification function (e.g., cross-entropy loss, accuracy) measuring performance or alignment. \\
        \midrule
        $\delta$ & The safety threshold indicating the maximum allowable expected value of $\phi$. \\
        \midrule
        $T \subseteq \mathbb{R}^p$ & A Locally Invariant Domain (LID). \\
        \midrule
        $|T|_S$ & A metric representing the size (e.g., volume) of the domain $T$. \\
        \midrule
        $\Pi_T$ & The projection operator that maps parameters onto the closure of $T$. \\
        \midrule
        $T(\alpha)$ & A parameterized abstract domain (e.g., orthotope) defined by parameters $\alpha$. \\
        \midrule
        $\mathcal{R}(\cdot)$ & A risk function that assigns a cost to individual parameters of a parametrized abstract domain to constrain the LID finding problem.\\
        \bottomrule
    \end{tabular}
\end{table}

\end{document}